\documentclass{article}
\usepackage{iclr2027_conference,times}

\usepackage{amsmath,amsfonts,bm}

\def\eqref#1{equation~\ref{#1}}

\def\1{\bm{1}}

\def\vq{{\bm{q}}}

\def\vs{{\bm{s}}}

\def\vu{{\bm{u}}}

\def\vx{{\bm{x}}}
\def\vy{{\bm{y}}}
\def\vz{{\bm{z}}}

\def\mE{{\bm{E}}}

\def\mV{{\bm{V}}}
\def\mW{{\bm{W}}}

\DeclareMathAlphabet{\mathsfit}{\encodingdefault}{\sfdefault}{m}{sl}
\SetMathAlphabet{\mathsfit}{bold}{\encodingdefault}{\sfdefault}{bx}{n}

\def\gB{{\mathcal{B}}}

\def\gS{{\mathcal{S}}}

\newcommand{\E}{\mathbb{E}}

\usepackage{hyperref}
\hypersetup{hypertexnames=false}
\usepackage{url}
\usepackage{graphicx}
\usepackage{capt-of}
\usepackage{wrapfig}
\usepackage{booktabs}
\usepackage{array}
\usepackage{algorithm}
\usepackage{algorithmic}
\usepackage{hypcap}
\usepackage{amssymb}
\usepackage{xcolor}
\usepackage{tikz}
\usetikzlibrary{arrows.meta,positioning,calc,fit}

\title{Invariant-Measure Reasoners:\\ Stable Representations for Latent Reasoning}

\author{Yuto Inui\\
The University of Osaka\\
\And
Takuya Konishi\\
The University of Osaka \&\\
RIKEN\\
\And
Yoshinobu Kawahara\\
The University of Osaka \&\\
RIKEN\\
\end{tabular}\hfil\linebreak[4]\hfil
\begin{tabular}[t]{l}
{\normalfont\texttt{\{yuto.inui,konishi,kawahara\}@ist.osaka-u.ac.jp}}
}

\newcommand{\ImR}{ImR}

\newcolumntype{L}[1]{>{\raggedright\arraybackslash}p{#1}}

\iclrfinalcopy
\begin{document}

\maketitle
\lhead{Invariant-Measure Reasoners: Stable Representations for Latent Reasoning}

\begin{abstract}
Latent reasoning models repeatedly update a latent state using the same recurrent block.
As the recurrent depth increases, the sequence of latent states may converge to a compact subset of the state space without necessarily converging to a fixed point.
Existing models typically predict by applying a prediction head to a single latent state.
However, the latent state can continue to change even after many updates, potentially making predictions unstable across recurrent depths.
To address this instability, we introduce invariant-measure reasoners (ImR), a framework that uses an invariant measure as a stable representation.
This measure describes the long-run distribution of latent states on the compact subset and is invariant under updates by the recurrent block.
ImR predicts from the expectation of the prediction head's output under this measure.
We use ImR in two ways: fine-tuning only the prediction head of existing models and training models from scratch.
Both approaches reduce prediction instability and improve accuracy in many settings on maze and Sudoku tasks.
In some settings, models trained with ImR exhibit non-fixed-point behavior more frequently than existing models yet achieve high accuracy even with such behavior, unlike existing models.
These results suggest that ImR can leverage otherwise destabilizing dynamics for latent reasoning.
\end{abstract}

\section{Introduction}
\label{sec:introduction}

Reasoning has attracted increasing attention in artificial intelligence as a capability to derive solutions through a sequence of logical steps rather than just recalling memorized answers~\citep{guo2025deepseek}.
Latent reasoning offers an alternative to chain of thought~\citep{wei2022chain} by performing intermediate computations in latent space without expressing each step in natural language~\citep{hao2025training}.
In particular, latent reasoning models that repeatedly apply the same recurrent block to a latent state have shown promise in complex structured reasoning tasks such as solving mazes and Sudoku puzzles~\citep{wang2025hierarchical,jolicoeur-martineau2025less}.
Increasing recurrent depth, defined as the number of recurrent updates, allows these models to devote more computation to challenging problems at test time~\citep{graves2017adaptive,schwarzschild2021can,geiping2025scaling}.
However, predictive performance can deteriorate beyond the recurrent depth used during training, a phenomenon known as overthinking~\citep{bansal2022endtoend}.
Recent advances, including input injection~\citep{bai2019deep,bansal2022endtoend} and deep supervision~\citep{wang2025hierarchical}, have mitigated this degradation and enabled effective test-time scaling~\citep{huang2026equilibrium,knutson2026logical}.

\begin{figure}[t]
\capstart
\begin{center}
\includegraphics[width=\linewidth]{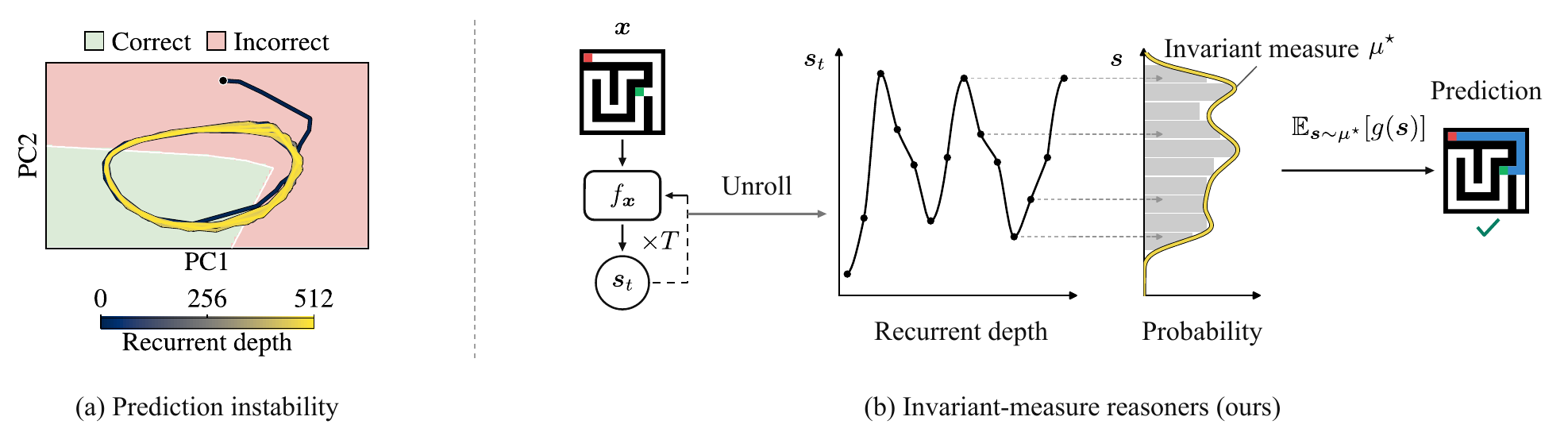}
\end{center}
\caption{(a) An example of prediction instability in equilibrium reasoners (EqR)~\citep{huang2026equilibrium} on a maze problem with the latent trajectory projected onto the first two principal components.
Trajectory color indicates recurrent depth, and pale green and red regions approximately indicate correct and incorrect predictions, respectively.
(b) Schematic of \ImR{}.
For an input $\vx$, repeated applications of the recurrent block $f_{\vx}$ to the latent state $\vs_t$ generate a latent trajectory.
\ImR{} predicts from the expectation $\E_{\vs\sim\mu^\star}[g(\vs)]$ under an invariant measure $\mu^\star$.}
\label{fig:latent-reasoning-overview}
\end{figure}

Despite these advances, latent reasoning models can still exhibit prediction instability across recurrent depths.
For a fixed input, recurrent updates define a discrete-time dynamical system.
In recent latent reasoning models, the resulting trajectories have been observed to approach a compact subset of the state space without converging to a single fixed point~\citep{anil2022path,geiping2025scaling,huang2026equilibrium,knutson2026logical}.
These models typically predict by applying a prediction head to the final latent state, so continued changes in that state can lead to changes in the predicted solution.
Figure~\ref{fig:latent-reasoning-overview}(a) illustrates this issue with a latent trajectory from equilibrium reasoners (EqR)~\citep{huang2026equilibrium}.
The trajectory remains within a bounded region and repeatedly traces a loop reminiscent of a limit cycle.
Along this trajectory, predictions alternate between correct and incorrect.
One way to obtain stable predictions is to predict from a fixed point, a latent state invariant under recurrent updates~\citep{bai2019deep,movahedi2026fixedpoint}.
Yet the continued changes in latent states observed in recent models raise the question of whether predictions can be stabilized without requiring convergence to a fixed point.

We introduce invariant-measure reasoners (\ImR), a framework that uses an invariant measure as a stable representation of latent dynamics.
An invariant measure is a probability distribution over latent states that is preserved by recurrent updates~\citep{eckmann1985ergodic}.
Bounded dynamics can admit such a measure even when individual states do not converge to a fixed point.
As illustrated in Figure~\ref{fig:latent-reasoning-overview}(b), \ImR{} predicts from the expectation of the prediction head's output under this measure.
Because recurrent updates preserve the measure, they also leave this expected output unchanged.
Prediction from a fixed point is recovered as the special case in which all probability is concentrated at that point.
In practice, \ImR{} approximates this ideal prediction by averaging the prediction head's outputs over a finite latent trajectory, using the corresponding empirical measure as an approximation to an invariant measure.
\ImR{} retains the existing model architecture, including the recurrent block and prediction head, and introduces no additional trainable parameters.

Our main contributions are as follows:
\begin{itemize}
    \item We introduce ImR, a framework for latent reasoning that uses invariant measures as a stable representation and includes prediction from a fixed point as a special case.

    \item We present two ways to use \ImR{}: fine-tuning only the prediction head of existing models and training models from scratch.
    Both approaches reduce prediction instability and improve accuracy in many settings on maze and Sudoku tasks.

    \item We show that, in some settings, models trained with \ImR{} exhibit non-fixed-point behavior more frequently than models trained to predict from the final state, yet form stable predictions and achieve high accuracy even with such behavior.
\end{itemize}

\section{Preliminaries}
\label{sec:preliminaries}
We first overview latent reasoning models, which can be seen as discrete-time dynamical systems.
To make the formulation concrete, we use structured reasoning tasks such as Sudoku and maze solving as running examples, but our formulation applies more broadly.

Structured reasoning tasks can typically be formulated as learning a mapping from an input configuration to a solution, both represented over a finite set of grid cells.
For a task with $n$ grid cells indexed in a fixed order, let $\vx\in\{1,\ldots,I\}^{n}$ denote the input configuration, where each cell takes one of the $I$ possible values.
We represent the solution as $\vy\in\{1,\ldots,C\}^{n}$, where $C$ is the number of output classes per cell.
For mazes, $\vx$ specifies the walls and the start and goal cells, while $\vy$ encodes the solution path.
For Sudoku, $\vx$ is a partially filled $9\times9$ grid, and $\vy$ represents a completed grid in which each row, column, and $3\times3$ box contains the digits 1 through 9 exactly once.

The latent reasoning models considered here repeatedly update a latent state that encodes an intermediate representation of the entire grid~\citep{bansal2022endtoend,huang2026equilibrium}.
For $t\in\mathbb{N}_0=\{0,1,2,\ldots\}$, we denote the latent state after $t$ recurrent updates by $\vs_t\in\gS\subseteq\mathbb{R}^{d}$, where $\gS$ is the state space and $d\in\mathbb{N}=\{1,2,\ldots\}$ is the state dimension.
We refer to $t$ as the recurrent depth.
For a fixed input $\vx$, these updates are performed by a recurrent block, which we represent as a continuous deterministic map $f_{\vx}:\gS\to\gS$.
This defines a discrete-time dynamical system:
\begin{equation}
\vs_{t+1}=f_{\vx}(\vs_t).
\label{eq:fixed-input-dynamics}
\end{equation}
The recurrent block $f_{\vx}$ is implemented by a neural network, such as a transformer or a convolutional neural network (CNN), with parameters shared across recurrent steps and $\vx$ supplied as input at each step.
The initial state $\vs_0$ may be computed from $\vx$~\citep{schwarzschild2021can,bansal2022endtoend} or sampled from an initialization distribution $\rho_0$ on $\gS$ that does not depend on $\vx$~\citep{wang2025hierarchical,huang2026equilibrium}.
For brevity, we write $f$ for $f_{\vx}$ when the input is clear from context.
Some models introduce noise into their recurrent updates, yielding stochastic latent dynamics~\citep{huang2026equilibrium,baek2026generative,sghaier2026probabilistic}.
Appendix~\ref{app:stochastic} presents the corresponding stochastic formulation.

To produce a solution, we stop the recurrent updates at depth $T\in\mathbb{N}$, which we refer to as the stopping depth.
Starting from $\vs_0$, these updates generate a finite sequence of latent states $\vs_1,\ldots,\vs_T$, which we call a finite latent trajectory and denote by $(\vs_t)_{t=1}^{T}$.
A prediction head $g:\gS\to\mathbb{R}^{n\times C}$, also implemented by a neural network, maps a latent state $\vs_t$ to a matrix of logits $\vz_t\in\mathbb{R}^{n\times C}$ over $C$ classes for each of the $n$ grid cells:
\begin{equation}
\vz_t=g(\vs_t).
\label{eq:instantaneous-logits}
\end{equation}
Prediction from the logits $\vz_T=g(\vs_T)$ at stopping depth $T$ yields the solution $\vy=(y_1,\ldots,y_n)^\top$ with $y_i=\arg\max_{c\in{1,\ldots,C}}z_{T,i,c}$, where $z_{T,i,c}\in\mathbb{R}$ denotes the logit for class $c$ at cell $i$ in $\vz_T$.
A notable feature of latent reasoning models is that the stopping depth $T$ can vary during training and testing, but the resulting solution can depend on its choice.

\section{Invariant-Measure Reasoners}
\label{sec:method}

As described in Section~\ref{sec:preliminaries}, prediction from the logits $\vz_T=g(\vs_T)$ at stopping depth $T$ can depend on the choice of $T$. Supporting different stopping depths is an important requirement, particularly for test-time scaling, where models perform more recurrent updates at test time than during training to improve generalization~\citep{schwarzschild2021can,geiping2025scaling,huang2026equilibrium,knutson2026logical}. However, latent trajectories in advanced models may converge to a compact subset of the state space without converging to a fixed point~\citep{anil2022path,geiping2025scaling,huang2026equilibrium,knutson2026logical}. Such dynamics may support reasoning beyond fixed-point convergence, but can also destabilize predictions: Figure~\ref{fig:latent-reasoning-overview}(a) shows a trajectory repeatedly crossing between regions associated with correct and incorrect solutions.

To address this prediction instability, we propose invariant-measure reasoners (\ImR), which use an invariant measure as a stable representation of latent dynamics and predict from the expected output of the prediction head under this measure. We first introduce invariant measures, then describe how \ImR{} forms predictions and approximates them using finite latent trajectories. We then present procedures for fine-tuning only the prediction head and training models from scratch.

\subsection{Invariant measures of latent dynamics}
\label{sec:attractors-measures}

We characterize the long-term behavior of latent trajectories in a latent reasoning model from a distributional perspective.
For a fixed input $\vx$, we consider the empirical measure $\hat{\mu}_T$ of the finite latent trajectory $(\vs_t)_{t=1}^{T}$ and assume that this measure converges weakly to a probability measure $\mu^\star$ on $\gS$ as $T\to\infty$:
\begin{equation}
    \hat{\mu}_T
    = \frac{1}{T}\sum_{t=1}^{T}\delta_{\vs_t}
    \Rightarrow \mu^\star,
    \label{eq:single-trajectory-measure}
\end{equation}
where $\delta_{\vs_t}$ is the Dirac measure assigning probability one to $\vs_t$, and $\Rightarrow$ denotes weak convergence.
The empirical measure assigns equal weight to each visit along the finite latent trajectory and describes a distribution of latent states even when the recurrent updates are deterministic.

The probability measure $\mu^\star$ is invariant under recurrent updates.
To express this property, we use the pushforward operator $f_\#$, also called the Perron--Frobenius operator~\citep{lasota1994chaos}.
For a probability measure $\mu$, $f_\#\mu$ is the distribution obtained by applying $f$ to states distributed according to $\mu$.
See Appendix~\ref{app:operator} for the formal definition of the pushforward operator.
Because $f$ is continuous, the weak convergence of the empirical measures along the trajectory implies
\begin{equation}
    f_\#\mu^\star=\mu^\star.
    \label{eq:invariant-measure}
\end{equation}
Such a measure is called an invariant measure~\citep{eckmann1985ergodic}, a fixed point of $f_\#$ in the space of probability measures.
It is preserved by recurrent updates even when individual latent states continue to evolve.

The support of $\mu^\star$ reflects the long-term behavior of the
latent trajectory.
If the trajectory converges to a compact set $\gB\subseteq\gS$,
in the sense that $\inf_{\vs\in\gB}\|\vs_t-\vs\|_2\to0$ as
$t\to\infty$, then $\mu^\star$ is supported on $\gB$, meaning that
$\mu^\star(\gB)=1$.
If the trajectory converges to a fixed point $\vs^\star$, we can
take $\gB=\{\vs^\star\}$, with $\mu^\star=\delta_{\vs^\star}$.
If it converges to a periodic orbit with $P$ distinct states, we
can take $\gB=\{\vs_1^\star,\ldots,\vs_P^\star\}$, with
$\mu^\star=P^{-1}\sum_{j=1}^{P}\delta_{\vs_j^\star}$.
More complex long-term behavior is also covered, provided that
the empirical measure converges.

Note that the invariant measure $\mu^\star$ may depend on the initial state $\vs_0$.
Convergence to the same measure from different initial states is an
additional desirable property, as in path-independent equilibrium
models~\citep{anil2022path}, but is not required by our convergence assumption.
Appendix~\ref{app:measure-convergence} discusses this distinction.

\subsection{Prediction from an invariant measure}
\label{sec:atr-prediction}

Building on the preceding formulation, we define \ImR{} as a framework for prediction based on the invariant measure $\mu^\star$ of latent dynamics. \ImR{} predicts from the expectation of the logits under $\mu^\star$:
\begin{equation}
    \vz^\star
    =
    \E_{\vs\sim\mu^\star}\!\left[g(\vs)\right]
    =
    \int_{\gS}g(\vs)\,d\mu^\star(\vs).
    \label{eq:atr-asymptotic-logits}
\end{equation}
Here, we assume that the prediction head $g$ is continuous on $\gS$ and that the logits $g(\vs_t)$ remain bounded along the latent trajectory.
Together with the weak convergence assumed in Section~\ref{sec:attractors-measures}, these conditions ensure that the expected logits $\vz^\star$ are finite.
By equation~\ref{eq:invariant-measure}, they are invariant under recurrent updates, since $\E_{\vs\sim\mu^\star}[g(f(\vs))]=\E_{\vs\sim f_\#\mu^\star}[g(\vs)]=\vz^\star$.
This does not require the prediction head itself to be invariant: $g(f(\vs))$ may differ from $g(\vs)$.

The invariant measure $\mu^\star$ describes limiting behavior and is generally not available explicitly. We approximate the expected logits $\vz^\star$ using the empirical measure $\hat{\mu}_T$ at a finite stopping depth $T$:
\begin{equation}
    \hat{\vz}_T
    =
    \E_{\vs\sim\hat{\mu}_T}\!\left[g(\vs)\right]
    =
    \frac{1}{T}\sum_{t=1}^{T}g(\vs_t).
    \label{eq:finite-atr-logits}
\end{equation}
As in Section~\ref{sec:preliminaries}, we select the class with the largest averaged logit at each grid cell to obtain the solution.
The weak convergence in equation~\ref{eq:single-trajectory-measure}, together with the conditions on $g$, implies $\hat{\vz}_T\to\vz^\star$ as $T\to\infty$.
If $\vz^\star$ has a unique maximum at each cell, the solution is constant for all sufficiently large $T$.

\ImR{} requires only a simple change in how latent reasoning models compute predictions.
Standard prediction from the final state uses $\vz_T=g(\vs_T)$, whereas \ImR{} uses the averaged logits $\hat{\vz}_T$.
The former can continue to fluctuate as latent states evolve, while the latter converge under the stated assumptions, even when individual states do not.
At finite depths, $\hat{\vz}_T$ can still depend on the stopping depth $T$, but averaging can reduce this sensitivity and thereby improve prediction robustness across recurrent depths.
\ImR{} also retains the existing recurrent block and prediction head and introduces no additional trainable parameters.

At finite recurrent depths, states from the initial transient can affect
the empirical approximation of $\mu^\star$.
To reduce their influence, one can discard the states from the first $B$
recurrent updates, with $0\leq B<T$, and average the logits over steps
$B+1,\ldots,T$.
For fixed $B$, the resulting empirical measure converges to the same
invariant measure $\mu^\star$ under the assumption in
Section~\ref{sec:attractors-measures}.
Appendix~\ref{app:finite-estimation} defines this estimator and gives
the choices used in our experiments.

\ImR{} also applies when the prediction head uses only part of the latent state.
In this case, the expected logits depend only on the corresponding marginal of $\mu^\star$, which is preserved when the full latent state is updated by $f$.
Appendix~\ref{app:models} specifies the latent state and the part used by the prediction head for each model in our experiments.

\subsection{Training with Prediction from an Invariant Measure}
\label{sec:atr-training}

We propose two approaches to training with \ImR{}: fine-tuning a pretrained model and training a model from scratch.
For fine-tuning, we retain the latent dynamics of the pretrained recurrent block $f$ and update only the prediction head $g$.
We compute a loss from the averaged logits in equation~\ref{eq:finite-atr-logits} and the target solution, treating the latent states as constants so that gradients pass only through $g$.
This approach aims to reduce prediction instability in an existing model without backpropagating through recurrent updates.
In contrast, training from scratch jointly optimizes $f$ and $g$ using a loss computed from the averaged logits, allowing the model to learn latent dynamics suited to prediction with \ImR{}.

For training from scratch, we combine \ImR{} with a deep supervision procedure~\citep{wang2025hierarchical,huang2026equilibrium} that repeatedly supervises predictions as recurrent computation continues. The procedure consists of an inner loop and an outer loop. Each inner loop performs $L$ recurrent updates with fixed model parameters and computes the averaged logits $\hat{\vz}_L$ over the resulting $L$ latent states. At each outer step, we compute the loss from these averaged logits 
\begin{wrapfloat}{algorithm}{r}{0.47\textwidth}
\begin{minipage}{\linewidth}
\setlength{\intextsep}{0pt}
\renewcommand{\algorithmiccomment}[1]{\hfill$\triangleright$\,#1}
\renewcommand{\algorithmicdo}{}
\begin{algorithm}[H]
\caption{Deep supervision for \ImR{}}
\label{alg:imr-training-overview}
\begin{algorithmic}[1]
\REQUIRE input $\vx$, target $\vy$
\STATE Initialize the latent state $\vs$
\FOR[outer loop]{each outer step}
    \STATE $\vz_{\mathrm{sum}}\gets 0$
    \FOR[inner loop]{$j=1,\ldots,L$}
        \IF{$j\leq L-K$}
            \STATE $\vs\gets\operatorname{NoGrad}[f_{\vx}(\vs)]$
        \ELSE
            \STATE $\vs\gets f_{\vx}(\vs)$
        \ENDIF
        \STATE $\vz_{\mathrm{sum}}\gets\vz_{\mathrm{sum}}+g(\vs)$
    \ENDFOR
    \STATE $\hat{\vz}_L\gets\vz_{\mathrm{sum}}/L$
    \COMMENT{empirical expectation}
    \STATE Compute the loss from $\hat{\vz}_L$ and $\vy$
    \STATE Backpropagate the loss
    \STATE Update the model parameters
    \STATE Detach $\vs$ for the next outer step
\ENDFOR
\end{algorithmic}
\end{algorithm}
\end{minipage}
\end{wrapfloat}
and the target solution, then update the model parameters. We retain the original model's loss function,
replacing prediction from the final state with prediction based on the averaged logits. The input remains fixed across outer steps, and each subsequent inner loop starts from the preceding inner loop's final latent state, detached from its computation graph.

To limit the cost of gradient computation, we use truncated backpropagation through time (TBPTT)~\citep{williams1990efficient}, retaining gradients through the recurrent block $f$ only for the final $K\leq L$ updates of each inner loop. All $L$ states contribute to training the prediction head $g$, while earlier states are treated as constants when computing gradients. We construct a new empirical measure in each inner loop and do not carry it over to the next. Each empirical measure therefore contains only states generated with the same model parameters. Algorithm~\ref{alg:imr-training-overview} summarizes the procedure, with further training details in Appendix~\ref{app:eqr-segmented-training}.

\section{Related Work}
\label{sec:related-work}

\paragraph{Equilibrium models and stability.}
Deep equilibrium models define latent representations as fixed points
of recurrent transformations~\citep{bai2019deep}.
To encourage convergence to a fixed point, Jacobian regularization has
been used in equilibrium and looped
models~\citep{bai2021stabilizing,yang2026stabilizing}.
However, such regularization can restrict the model class and does not
guarantee convergence to a fixed point~\citep{bai2021stabilizing}.
Recent studies have also reported long-term behavior beyond convergence to a fixed point in latent reasoning models~\citep{anil2022path,geiping2025scaling,huang2026equilibrium,knutson2026logical}.
In this paper, we consider latent trajectories whose empirical measures converge weakly, encompassing both convergence to a fixed point and more general long-term behavior.
\ImR{} generalizes the formulation based on convergence to a fixed point in the state space
to one based on weak convergence of empirical measures, where the representation satisfies $f_\#\mu^\star=\mu^\star$.

\paragraph{Adaptive computation and halting.}
Adaptive computation time (ACT) learns a halting signal that determines
the stopping depth and weights for averaging intermediate states and
outputs~\citep{graves2017adaptive}.
PonderNet learns a distribution over stopping depths and samples a
stopping depth for prediction at inference~\citep{banino2021pondernet}.
\ImR{} predicts from the expectation of the prediction head's output
under the empirical measure of a latent trajectory.
Its weights reflect state visit frequencies rather than learned halting
probabilities, and it introduces no additional trainable parameters.
\ImR{} remains complementary to halting mechanisms, which can determine
when to stop collecting states to estimate the invariant measure.

\section{Experiments}
\label{sec:experiments}

We conducted numerical experiments to evaluate whether \ImR{}
reduces prediction instability and improves accuracy across
different architectures.
We used EqR~\citep{huang2026equilibrium}
and the deep thinking network with progressive training and
recall~\citep{bansal2022endtoend}, which we refer to as DT.
For both architectures, we fine-tuned only the prediction head.
With EqR, we also trained models from scratch and assessed
prediction stability separately for trajectories classified as
fixed and non-fixed.
With DT, we examined prediction instability on out-of-distribution
(OOD) mazes.

\paragraph{Model architectures.}
EqR uses transformer blocks on Maze-Unique and multilayer perceptrons
(MLPs) to mix information across positions and channels on
Sudoku-Extreme, while DT uses a CNN.
\ImR{} models use the EqR architecture on Maze-Unique and
Sudoku-Extreme and the DT architecture on OOD mazes.

EqR adopts hierarchical iteration, which uses a high-level latent
state $\vs^{(H)}$ and a low-level latent state $\vs^{(L)}$.
Together, these states form the full latent state $\vs$.
At each recurrent step, the same module updates $\vs^{(L)}$ several
times and then $\vs^{(H)}$ once.
EqR also uses a puzzle prefix to extend the input representation
and both latent states with additional positions beyond those
corresponding to grid cells.
In the input representation, the prefix consists of a learned
vector shared across problems and zero vectors.
The corresponding positions in both latent states are updated
along with the grid positions, providing additional dimensions
for recurrent computation.
The prediction head uses only the grid positions of $\vs^{(H)}$.
Appendix~\ref{app:models} gives architectural details.

\paragraph{Baselines.}
We included the released EqR and DT models as baselines without
fine-tuning.
To distinguish the effect of \ImR{} on prediction stability
from that of fine-tuning the prediction head, we also included
fine-tuned baselines, EqR (FT) and DT (FT).
These baselines were trained using each model's original task
loss applied to the logits $g(\vs_T)$ from the final latent state.
We obtained \ImR{} (FT) by fine-tuning with the same loss
applied to the expectation of $g$ under the empirical measure
of the latent trajectory.
For each architecture, both fine-tuning methods started from
the same released weights.

For the EqR experiments without the puzzle prefix, we also trained
EqR baselines from scratch using prediction from the final latent
state.
In the EqR experiments, \ImR{} without the suffix ``(FT)''
denotes models trained from scratch.
We used 3 training seeds for each configuration trained or
fine-tuned in this study.
Results for each released baseline are from a single pretrained model.

\paragraph{Evaluation.}
At each recurrent depth $t$, EqR, DT, and their fine-tuned versions
predict from the current latent state.
All \ImR{} models predict from the expectation of $g$ under the
empirical measure $\hat{\mu}_t$, which uses all states from depths
$1,\ldots,t$.
We plot accuracy as a function of recurrent depth.
Accuracy alone does not fully capture prediction instability:
changes between correct and incorrect predictions can offset
each other across instances, and changes between different
incorrect predictions do not affect accuracy.

We also assess prediction stability for each instance over an
interval specified in each experiment.
An instance is stable correct if the model outputs the same
correct prediction throughout the interval.
It is stable incorrect if the model outputs the same incorrect
prediction throughout the interval.
It is unstable if its prediction changes at least once,
including changes between different incorrect predictions.

\begin{figure}[t]
\capstart
\begin{center}
\includegraphics[width=\linewidth,trim=0bp 4bp 0bp 1bp,clip]{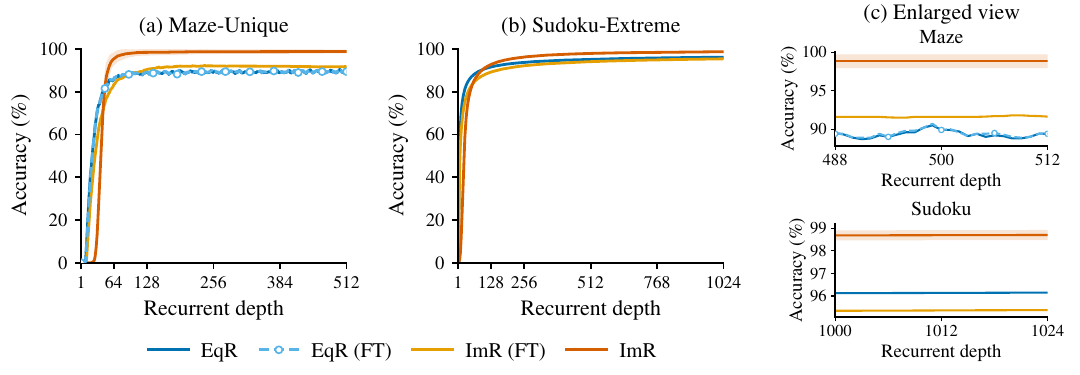}
\end{center}
\caption{Accuracy across recurrent depth on Maze-Unique (a) and
Sudoku-Extreme (b), with enlarged views (c) over the final 25 recurrent depths.
For both fine-tuned models and \ImR{}, curves show
means and shaded regions show standard deviation across
3 training seeds.}
\label{fig:eqr-main}
\end{figure}

\subsection{EqR on Maze-Unique and Sudoku-Extreme}
\label{sec:exp-eqr}

We evaluated EqR and \ImR{} on
Maze-Unique~\citep{huang2026equilibrium}
and Sudoku-Extreme~\citep{wang2025hierarchical}.
Maze-Unique requires marking the unique path from the start cell
to the goal cell on a $30\times30$ grid without crossing walls.
Sudoku-Extreme requires completing a partially filled
$9\times9$ Sudoku grid.
For both tasks, a prediction is correct only if the entire predicted grid matches
the target.
We used the released datasets and constructed validation
sets for hyperparameter selection.

We largely followed the model configurations, datasets, and hyperparameters
of EqR~\citep{huang2026equilibrium}.
Accuracy and prediction stability were evaluated up to recurrent
depths of $T=512$ on Maze-Unique and $T=1024$ on Sudoku-Extreme,
with transition noise set to zero during evaluation.
Appendix~\ref{app:experimental-details} provides the data splits
and detailed training and evaluation settings.

We compared models with and without the puzzle prefix to examine
how latent state dimensionality relates to learned
dynamics and prediction stability.
All models without the prefix were trained from scratch.
For each model, the prefix configuration remained unchanged
throughout training and evaluation.

To assess prediction stability under different latent dynamics,
we classified trajectories as fixed or non-fixed using the relative
$L_\infty$ residual between successive values of $\vs^{(H)}$ at the grid positions.
A trajectory was classified as \emph{fixed} if this residual remained
below $0.05$ throughout the final 64 recurrent updates, and as
\emph{non-fixed} otherwise.
These labels describe state changes over the evaluated interval
and do not guarantee convergence of the full latent state to a
fixed point.
Within each group, we evaluated prediction stability over the
final 64 recurrent depths.
For configurations with multiple training seeds, we computed the
proportions separately for each trained model and reported their
mean and standard deviation.
Appendix~\ref{app:eqr-fixed-diagnostics} provides the residual
definition and calculation details.

\paragraph{Results.}
With the puzzle prefix, \ImR{} trained from scratch achieves
higher accuracy than the released EqR model on both tasks at large
recurrent depths, as shown in Figure~\ref{fig:eqr-main}.
At these depths, \ImR{} (FT) achieves higher accuracy than EqR (FT)
on Maze-Unique but lower accuracy than the released EqR model
on Sudoku-Extreme.
For Sudoku-Extreme, EqR (FT) is not shown separately because
fine-tuning did not improve validation accuracy and the initial
prediction head was retained.
Table~\ref{tab:final-depth-accuracy} in Appendix~\ref{app:final-depth-accuracy}
reports the corresponding accuracies at the final recurrent depth.
The larger gains from training from scratch suggest that jointly
training the recurrent block and prediction head with \ImR{} can
be more effective than fine-tuning the prediction head alone.

Table~\ref{tab:eqr-from-scratch-dynamics} shows that \ImR{} reduces
prediction instability in the non-fixed group.
Predictions in the fixed group are predominantly stable correct
across all models, whereas EqR predictions in the non-fixed group
are mostly unstable.
\ImR{} (FT) reduces this instability on both tasks while leaving
the recurrent block unchanged, showing that prediction stability
can improve without altering the latent dynamics.
On Sudoku-Extreme, however, most non-fixed predictions under
\ImR{} (FT) are stable incorrect, so this stabilization does not
translate into higher accuracy.
Among \ImR{} models trained from scratch, those without the prefix
have higher mean non-fixed rates than those with it, reaching
about 50\% on both tasks, with substantial variation across seeds.
Nevertheless, over 97\% of predictions in their non-fixed group
are stable correct on average.
These results show that \ImR{} can improve prediction stability
and accuracy even for trajectories classified as non-fixed.

Figure~\ref{fig:eqr-nonfixed-pca} in Appendix~\ref{app:eqr-pca} shows principal component analysis projections
of additional latent trajectories.

\begin{table}[t]
\caption{Trajectory groups and prediction stability.
Entries with $\pm$ report mean and standard deviation across
3 training seeds.
EqR$^\dagger$ reports results for both EqR and EqR (FT),
which match on all reported metrics.
For each dataset and state dimension of $\vs^{(H)}$, bold marks the highest stable
correct proportion and the lowest unstable proportion in the non-fixed group.}
\label{tab:eqr-from-scratch-dynamics}

\setlength{\tabcolsep}{2.7pt}
\begin{center}
\begin{tabular}{@{}l*{7}{r}@{}}
\toprule
& & \multicolumn{3}{c}{Fixed}
& \multicolumn{3}{c}{Non-fixed} \\
\cmidrule(lr){3-5}\cmidrule(l){6-8}
Model
& \shortstack{Non-fixed\\rate}
& \shortstack{Stable\\correct}
& \shortstack{Stable\\incorrect}
& Unstable
& \shortstack{Stable\\correct}
& \shortstack{Stable\\incorrect}
& Unstable \\
\midrule

\multicolumn{8}{@{}l}{\textit{Sudoku-Extreme}: dimension of $\vs^{(H)}$:  $49.7\mathrm{K}$; parameters $5.03\mathrm{M}$} \\
EqR$^\dagger$ & $4.1$ & $100.0$ & $0.0$ & $0.0$ & $3.6$ & $0.0$ & $96.4$ \\
\ImR{} (FT) & $4.1 \pm 0.0$ & $99.2 \pm 0.0$ & $0.0 \pm 0.0$ & $0.8 \pm 0.0$ & $3.4 \pm 0.0$ & $79.7 \pm 0.1$ & $\bm{16.9 \pm 0.1}$ \\
\ImR{} & $1.2 \pm 0.2$ & $99.7 \pm 0.1$ & $0.0 \pm 0.0$ & $0.2 \pm 0.1$ & $\bm{8.0 \pm 12.2}$ & $44.5 \pm 5.8$ & $47.5 \pm 8.0$ \\
\addlinespace[2pt]
\multicolumn{8}{@{}l}{\textit{Sudoku-Extreme}: dimension of $\vs^{(H)}$: $41.5\mathrm{K}$; parameters $4.86\mathrm{M}$} \\
EqR & $4.5 \pm 0.9$ & $100.0 \pm 0.0$ & $0.0 \pm 0.0$ & $0.0 \pm 0.0$ & $11.6 \pm 18.9$ & $0.0 \pm 0.0$ & $88.4 \pm 18.9$ \\
\ImR{} & $50.3 \pm 33.7$ & $99.7 \pm 0.1$ & $0.0 \pm 0.0$ & $0.3 \pm 0.1$ & $\bm{97.9 \pm 0.8}$ & $0.8 \pm 0.3$ & $\bm{1.4 \pm 0.5}$ \\
\midrule

\multicolumn{8}{@{}l}{\textit{Maze-Unique}: dimension of $\vs^{(H)}$: $117.2\mathrm{K}$; parameters $264.1\mathrm{K}$} \\
EqR$^\dagger$ & $14.7$ & $98.5$ & $1.5$ & $0.0$ & $16.3$ & $0.7$ & $83.0$ \\
\ImR{} (FT) & $14.7 \pm 0.0$ & $98.4 \pm 0.0$ & $1.1 \pm 0.0$ & $0.6 \pm 0.0$ & $51.7 \pm 0.0$ & $23.8 \pm 1.8$ & $24.5 \pm 1.8$ \\
\ImR{} & $3.6 \pm 3.2$ & $99.8 \pm 0.2$ & $0.2 \pm 0.1$ & $0.1 \pm 0.1$ & $\bm{71.9 \pm 4.5}$ & $24.9 \pm 8.4$ & $\bm{3.2 \pm 5.5}$ \\
\addlinespace[2pt]
\multicolumn{8}{@{}l}{\textit{Maze-Unique}: dimension of $\vs^{(H)}$: $115.2\mathrm{K}$; parameters $263.9\mathrm{K}$} \\
EqR & $18.6 \pm 21.2$ & $99.9 \pm 0.1$ & $0.1 \pm 0.1$ & $0.0 \pm 0.0$ & $11.7 \pm 17.6$ & $0.2 \pm 0.4$ & $88.1 \pm 17.4$ \\
\ImR{} & $50.8 \pm 28.4$ & $100.0 \pm 0.1$ & $0.0 \pm 0.1$ & $0.0 \pm 0.0$ & $\bm{97.5 \pm 3.9}$ & $2.1 \pm 3.3$ & $\bm{0.5 \pm 0.6}$ \\

\bottomrule
\end{tabular}
\end{center}
\end{table}

\subsection{DT on OOD Mazes}
\label{sec:exp-dt}

Prediction instability may increase when a pretrained recurrent model
is applied to more challenging problems.
We examined whether fine-tuning with \ImR{} can reduce this instability
on OOD mazes.
We characterized maze structure by the degree of the start cell,
defined as its number of traversable neighbors.
The released DT model was pretrained on $9\times9$ mazes whose
start cell had degree $1$.
We refer to mazes satisfying these conditions as in-distribution
(ID) mazes.

Following the OOD maze setting of \citet{knutson2026logical}, we evaluated
mazes whose start cells had degrees from $1$ to $4$.
The test set contained 100 mazes at each of 10 sizes from
$9\times9$ to $99\times99$, for $1{,}000$ mazes in total.
Mazes with a start cell of degree greater than $1$ are OOD relative
to the pretraining conditions, and larger mazes are also OOD in size.
We analyzed the results separately by the degree of the start cell.
Figure~\ref{fig:dt-maze-examples} in Appendix~\ref{app:datasets-validation} shows examples of ID and OOD inputs.

We compared the released DT model with DT (FT) and
\ImR{} (FT).
Fine-tuning and hyperparameter selection used only ID mazes.
We evaluated predictions at every recurrent depth from $1$ through
$5{,}000$ and assessed prediction stability over depths
$4{,}900$--$5{,}000$.
Appendix~\ref{app:experimental-details} describes the dataset construction and gives detailed
training and evaluation settings.

\begin{figure}[t]
\capstart
\begin{center}
\includegraphics[width=\linewidth,trim=0bp 4bp 0bp 1bp,clip]{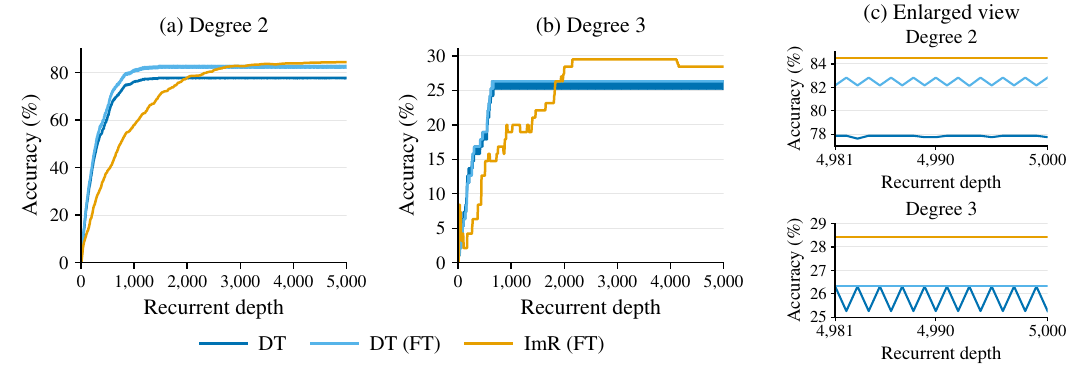}
\end{center}
\caption{Accuracy across recurrent depth on OOD mazes whose start
cell has degree 2 (a) or 3 (b), with enlarged views (c) over the final 20 recurrent depths.
For both fine-tuned models, curves show means and shaded regions show standard deviation across 3 training seeds.}
\label{fig:dt-maze-results}
\end{figure}

\paragraph{Results.}
The accuracy gains of \ImR{} (FT) emerge at larger recurrent depths,
as shown in Figure~\ref{fig:dt-maze-results}.
For mazes whose start cell has degree 2 or 3, its accuracy initially
increases more slowly than that of the released DT model, but
eventually surpasses it and remains higher through the end of
evaluation.
At the final depth, \ImR{} (FT) also achieves higher accuracy than
DT (FT) in both groups
and across the full test set.
Table~\ref{tab:final-depth-accuracy} in Appendix~\ref{app:final-depth-accuracy}
reports the final-depth accuracies for the degree-2 and degree-3 groups.
These results indicate that additional recurrent computation can
improve predictions through the empirical expectation even after
accuracy from the final state has largely plateaued.
Figure~\ref{fig:dt-size-depth-accuracy} in Appendix~\ref{app:dt-full-results}
provides results separately for each maze size.

Table~\ref{tab:dt-prediction-stability} shows that these accuracy gains
are accompanied by more stable predictions for individual instances.
Relative to DT (FT), \ImR{} (FT) further reduces the proportion
of unstable predictions and increases the stable correct proportion
in both groups.
Although the stable incorrect proportion also increases,
particularly for degree 3, the gains over DT (FT) show that
\ImR{} can obtain more stable and accurate predictions from
the same pretrained dynamics.
These gains are achieved under the
\begin{wraptable}{r}{0.54\textwidth}
\vspace{-10.5pt}
\caption{Prediction stability on OOD mazes over depths
$4{,}900$--$5{,}000$.
Fine-tuned models report mean $\pm$ standard deviation across 3 training seeds.
For each degree of the start cell, bold marks the highest stable correct
proportion and the lowest unstable proportion.}
\label{tab:dt-prediction-stability}
\begin{center}
\setlength{\tabcolsep}{3pt}
\begin{tabular}{@{}lrrr@{}}
\toprule
Model & \shortstack{Stable\\correct (\%)}
& \shortstack{Stable\\incorrect (\%)} & Unstable (\%) \\
\midrule
\multicolumn{4}{@{}l}{\emph{Degree of the start cell: 2}} \\
DT & $76.6$ & $17.3$ & $6.1$ \\
DT (FT) & $81.5 \pm 0.1$ & $13.9 \pm 0.1$ & $4.5 \pm 0.1$ \\
\ImR{} (FT) & $\bm{84.5 \pm 0.0}$ & $15.1 \pm 0.1$ & $\bm{0.4 \pm 0.1}$ \\
\midrule
\multicolumn{4}{@{}l}{\emph{Degree of the start cell: 3}} \\
DT & $25.3$ & $57.9$ & $16.8$ \\
DT (FT) & $26.3 \pm 0.0$ & $62.8 \pm 0.6$ & $10.9 \pm 0.6$ \\
\ImR{} (FT) & $\bm{28.4 \pm 0.0}$ & $70.2 \pm 0.6$ & $\bm{1.4 \pm 0.6}$ \\
\bottomrule
\end{tabular}
\end{center}
\end{wraptable}
evaluated OOD conditions
using only ID mazes for fine-tuning and hyperparameter selection,
with the recurrent block unchanged.

For mazes whose start cell has degree 1, all models solve every
test maze at the final depth, including mazes larger than those
seen during pretraining.
Thus, the gains on degree-2 and degree-3 mazes are obtained while
preserving DT's performance on degree-1 mazes that are OOD in size.
The test set contains only one maze whose start cell has degree 4,
and none of the evaluated models produces a correct solution
for this instance.

\section{Discussion}
\label{sec:discussion}

The results without the puzzle prefix suggest that \ImR{} can
support accurate reasoning in lower-dimensional latent spaces.
Compared with \ImR{} models trained with the prefix, those trained
without it used fewer latent state dimensions and had a higher mean
non-fixed rate.
Nevertheless, their predictions in the non-fixed group were
predominantly stable correct.
Understanding how latent trajectories and their induced measures
represent solutions as latent state dimensionality varies could
guide the design of more efficient recurrent models.

Our experiments used fixed stopping depths.
Developing adaptive stopping rules that balance predictive accuracy
and computation remains an important direction for future work.

\section{Conclusion}
\label{sec:conclusion}

We introduced \ImR{} to address prediction instability at large
recurrent depths in latent reasoning models.
\ImR{} uses an invariant measure as a stable representation of
latent dynamics and predicts from the expectation of the prediction
head's output under this measure.
Experiments on maze and Sudoku tasks showed reduced prediction
instability and improved accuracy in many settings.
In particular, models trained with \ImR{} produced stable and
correct predictions even when their latent states continued to evolve.
These findings support a broader formulation of latent reasoning
based on convergence of empirical measures, allowing stable
prediction without requiring individual states to converge to a
fixed point.

\subsection*{AI use statement}

We used generative AI tools to assist with drafting and revising
the manuscript, searching for relevant literature, and discussing
mathematical formulations and their assumptions.
We also used these tools to write code for the numerical experiments and to discuss the interpretation and
presentation of experimental results.
The authors reviewed and revised the AI-assisted material, including the generated code, and
take full responsibility for the final content of this work.

\subsection*{Reproducibility statement}

Section~\ref{sec:method} describes the prediction rule and
training procedures for \ImR{}, with an overview of training
from scratch in Algorithm~\ref{alg:imr-training-overview}.
Appendix~\ref{app:datasets-validation} describes the datasets,
their preparation, and the training, validation, and test splits.
Appendix~\ref{app:models} specifies the model architectures,
while Appendix~\ref{app:finite-estimation} details the training
and evaluation settings and the procedures for selecting
hyperparameters and checkpoints.
Appendix~\ref{app:eqr-fixed-diagnostics} provides the criteria
for classifying latent trajectories and the procedures for
calculating prediction stability proportions within each group.
Appendix~\ref{app:latent-dynamics-visualization} describes how
the PCA projection and prediction regions in
Figure~\ref{fig:latent-reasoning-overview}(a) were constructed
and reports the corresponding predictions from the actual
latent states.

\subsubsection*{Acknowledgments}

This work was supported by JST BOOST Grant Number JPMJBS2402,
JSPS KAKENHI Grant Numbers JP25K15230 and JP26H02498, and JST Moonshot R\&D Program
Grant Number \mbox{JPMJMS263K}.

\bibliography{iclr2027_conference}
\bibliographystyle{iclr2027_conference}

\appendix

\section{An Operator-Theoretic Perspective on Latent Reasoning}
\label{app:operator-perspective}

This appendix describes how recurrent updates act on probability
measures and on functions of latent states.
Appendix~\ref{app:stochastic} introduces stochastic latent dynamics.
Appendix~\ref{app:operator} presents the pushforward and Koopman operators
and relates prediction under an invariant measure to the invariant
component of the prediction head.
Appendix~\ref{app:measure-convergence} distinguishes convergence along a
trajectory from convergence to a common measure across initializations.

\subsection{Stochastic Latent Dynamics}
\label{app:stochastic}

Some latent reasoning models, including EqR, sample noise during
their recurrent updates.
We collect all noise samples used within one complete recurrent
update into $\epsilon_t$.
For EqR, the full latent state contains both $\vs^{(H)}$ and $\vs^{(L)}$,
and one recurrent update consists of $J$ internal updates of
$\vs^{(L)}$ followed by one update of $\vs^{(H)}$.
Thus, we write
$\epsilon_t=(\epsilon_{t,1},\ldots,\epsilon_{t,J},\epsilon_{t,H})$,
where each entry is an independently sampled Gaussian noise vector
added during the corresponding internal update.
The stochastic extension of equation~\ref{eq:fixed-input-dynamics} is
\begin{equation}
    \vs_{t+1}
    =
    \tilde{f}_{\vx}(\vs_t,\epsilon_t),
    \qquad
    \epsilon_t
    \overset{\mathrm{i.i.d.}}{\sim}
    \nu=\mathcal{N}(0,\sigma^2 I).
    \label{eq:stochastic-recurrent-update}
\end{equation}
Given these noise samples, $\tilde{f}_{\vx}$ performs the internal
updates in sequence to produce the next latent state.
Appendix~\ref{app:models} gives the explicit update equations.
Setting all noise samples to zero recovers the deterministic map:
\begin{equation}
    \tilde{f}_{\vx}(\vs,0)=f_{\vx}(\vs).
    \label{eq:zero-noise-map}
\end{equation}
The input, model parameters, and noise distribution remain fixed
throughout the trajectory, and the noise sequence is independent
of the initial state.

\subsection{Pushforward and Koopman Operators}
\label{app:operator}

To describe how recurrent updates transform state distributions,
we use the pushforward introduced in the main text.
For a probability measure $\mu$ on $\gS$, $(f_{\vx})_\#\mu$
is the distribution obtained by applying $f_{\vx}$ to a state
sampled from $\mu$.
For every measurable set $A\subseteq\gS$,
\begin{equation}
    \bigl((f_{\vx})_\#\mu\bigr)(A)
    =
    \mu\!\left(f_{\vx}^{-1}(A)\right)
    =
    \mu\!\left(
        \left\{\vs\in\gS:f_{\vx}(\vs)\in A\right\}
    \right).
    \label{eq:pushforward-definition}
\end{equation}
Here, $f_{\vx}^{-1}(A)$ is the preimage of $A$, the set of
latent states that $f_{\vx}$ maps into $A$.

For fixed noise samples $\epsilon$, the same definition applies
to $\tilde{f}_{\vx}(\cdot,\epsilon)$.
The expectation of this pushforward under the noise distribution
$\nu$ defines the operator $\mathcal{P}$, also called
the Perron--Frobenius or transfer operator.
Its dual, the Koopman operator $\mathcal{K}$, acts on scalar
functions $\varphi$ of latent states, called observables~\citep{mezic2004comparison}:
\begin{equation}
    \begin{aligned}
        \mathcal{P}\mu
        &=
        \int
        \bigl(\tilde{f}_{\vx}(\cdot,\epsilon)\bigr)_\#\mu
        \,d\nu(\epsilon),\\
        (\mathcal{K}\varphi)(\vs)
        &=
        \int
        \varphi\!\left(\tilde{f}_{\vx}(\vs,\epsilon)\right)
        \,d\nu(\epsilon).
    \end{aligned}
    \label{eq:pf-koopman}
\end{equation}
The deterministic dynamics correspond to $\nu=\delta_0$,
which gives
\[
    \mathcal{P}\mu=(f_{\vx})_\#\mu,
    \qquad
    (\mathcal{K}\varphi)(\vs)
    =(\varphi\circ f_{\vx})(\vs)
    =\varphi(f_{\vx}(\vs)).
\]
The prediction head $g$ in the main text defines a scalar observable
$\vs\mapsto(g(\vs))_{i,c}$ for each grid cell $i$ and class $c$.

We write
$\langle\mu,\varphi\rangle
=\int_{\gS}\varphi(\vs)\,d\mu(\vs)$
for the pairing between a measure and an observable.
For bounded measurable observables, the definitions above give
\begin{equation}
    \begin{aligned}
        \langle\mathcal{P}\mu,\varphi\rangle
        &=
        \int_{\gS}\varphi(\vs)\,d(\mathcal{P}\mu)(\vs)\\
        &=
        \int\!\int_{\gS}
        \varphi\!\left(\tilde{f}_{\vx}(\vs,\epsilon)\right)
        \,d\mu(\vs)\,d\nu(\epsilon)\\
        &=
        \int_{\gS}
        \left[
            \int
            \varphi\!\left(\tilde{f}_{\vx}(\vs,\epsilon)\right)
            \,d\nu(\epsilon)
        \right]\,d\mu(\vs)\\
        &=
        \int_{\gS}(\mathcal{K}\varphi)(\vs)\,d\mu(\vs)
        =
        \langle\mu,\mathcal{K}\varphi\rangle.
    \end{aligned}
    \label{eq:operator-duality}
\end{equation}

If $\mu_t$ denotes the distribution of latent states at depth $t$,
then this distribution evolves according to
$\mu_{t+1}=\mathcal{P}\mu_t$.
The same recurrent computation therefore admits three descriptions:
the recurrent map evolves latent states,
$\mathcal{K}$ describes the expected evolution of functions
of those states, such as individual logits, and
$\mathcal{P}$ evolves the distribution of latent states.
Duality connects the latter two descriptions.
The expectation of an observable under the evolved distribution
equals the expectation of the evolved observable under
the original distribution.

An invariant measure $\mu$ satisfies $\mathcal{P}\mu=\mu$.
For stochastic dynamics, such a measure is also called
a stationary measure.
Duality then gives
\begin{equation}
    \langle\mu,\mathcal{K}\varphi\rangle
    =
    \langle\mathcal{P}\mu,\varphi\rangle
    =
    \langle\mu,\varphi\rangle.
    \label{eq:invariant-expectation}
\end{equation}
Thus, recurrent updates preserve expectations under an invariant
measure, even though individual latent states may change.

\paragraph{Eigenfunctions and Fourier averages.}
We now consider the deterministic dynamics in the main text.
An invariant measure is an eigenmeasure of $\mathcal{P}$
with eigenvalue $1$.
Likewise, a nonzero observable satisfying
$\mathcal{K}\varphi=\varphi$ is a Koopman eigenfunction
with eigenvalue $1$.
This equation expresses invariance of the observable under
recurrent updates, since $\varphi\circ f_{\vx}=\varphi$.

Applying $\mathcal{K}$ coordinatewise to the prediction head gives
$(\mathcal{K}^{t}g)(\vs_0)=g(\vs_t)$.
The empirical expectation in equation~\ref{eq:finite-atr-logits}
is therefore the finite Fourier average at frequency zero,
evaluated at $\vs_0$~\citep{mauroy2012use}.
The corresponding limiting function is
\begin{equation}
    \bar{g}(\vs)
    =
    \lim_{T\to\infty}
    \frac{1}{T}\sum_{t=1}^{T}
    (\mathcal{K}^{t}g)(\vs),
    \label{eq:koopman-fourier-average}
\end{equation}
wherever this limit exists.
Shifting the starting state by one recurrent update leaves
the limit unchanged, so
$\bar{g}(f_{\vx}(\vs))=\bar{g}(\vs)$, or equivalently
$\mathcal{K}\bar{g}=\bar{g}$.
Each nonzero coordinate function of $\bar{g}$ is therefore
a Koopman eigenfunction with eigenvalue $1$.

Under the empirical convergence assumption and the conditions
on $g$ in the main text, the prediction from the invariant
measure satisfies
\begin{equation}
    \vz^\star
    =
    \int_{\gS}g(\vs)\,d\mu^\star(\vs)
    =
    \bar{g}(\vs_0).
    \label{eq:invariant-head-function}
\end{equation}
This perspective clarifies that \ImR{} does not assume
$g\circ f_{\vx}=g$ or enforce this identity through training
or the construction of the prediction head.
It predicts using the invariant component $\bar{g}$ extracted
from $g$ by the Fourier average, even when
$g(f_{\vx}(\vs))\neq g(\vs)$.

\subsection{Dependence on Initialization}
\label{app:measure-convergence}

For the deterministic dynamics in the main text, different initial
states can lead to different long-term behavior even when the input
$\vx$ is fixed and the trajectories remain bounded.
The empirical convergence in
equation~\ref{eq:single-trajectory-measure} does not require
the invariant measure $\mu^\star$ to be the same across initial states.

We group initial states according to the invariant measure
to which their empirical measures converge.
Let $\{\mu_\alpha^\star\}_{\alpha\in\mathcal I}$ denote these
distinct invariant measures, where $\mathcal I$ is an index set.
For each $\alpha\in\mathcal I$, we define the corresponding
class of initial states as
\begin{equation}
    \gS_\alpha
    =
    \left\{
        \vs_0\in\gS :
        \hat{\mu}_T\to\mu_\alpha^\star
        \quad (T\to\infty)
    \right\}.
    \label{eq:ergodic-component-convergence}
\end{equation}
Here, $\hat{\mu}_T$ is the empirical measure in
equation~\ref{eq:single-trajectory-measure} for the trajectory
initialized at $\vs_0$, and convergence is weak.
The sets $\gS_\alpha$ are pairwise disjoint and partition
the initial states for which the empirical convergence holds.
When each limiting invariant measure $\mu_\alpha^\star$ is ergodic,
the partition $\{\gS_\alpha\}_{\alpha\in\mathcal I}$ corresponds to
an ergodic partition~\citep{mezic2004comparison}, although our
formulation does not assume ergodicity.

We do not assume that $\mu^\star$ is independent of initialization,
and the notation in the main text suppresses this dependence.
Independence from initialization is a desirable property that
may emerge through training.
However, for some inputs and initial states, the learned dynamics
may still lead trajectories toward subsets associated with
incorrect predictions, as illustrated by the analysis of
EqR~\citep{huang2026equilibrium}.

\section{Experimental and Training Details}
\label{app:experimental-details}

We first describe the datasets and model architectures used
in our experiments.
We then present the training and evaluation settings.
Finally, we describe how we classify latent trajectories
and visualize latent dynamics.

\subsection{Datasets and Validation}
\label{app:datasets-validation}

\paragraph{Datasets for EqR.}
We used the Maze-Unique and Sudoku-Extreme training and test sets
released with EqR~\citep{huang2026equilibrium}.\footnote{\url{https://huggingface.co/datasets/locuslab/EqR-data}}
Maze-Unique contains $30\times30$ mazes with unique solution paths
of 100--140 moves.
The target marks the solution path on the full grid.
For Sudoku-Extreme, the training set contains $1{,}000$ base puzzles
on $9\times9$ grids, each with $1{,}000$ augmented versions.
Augmentation used digit permutations, transposition, and row and
column permutations that preserve Sudoku constraints.
Including the original puzzles gives $1{,}001{,}000$ training examples.

We constructed validation sets for hyperparameter selection.
For Maze-Unique, we generated $1{,}000$ mazes under the same
conditions as the training set released with EqR and excluded
exact matches to mazes in that set.
For Sudoku-Extreme, we selected $10{,}000$ unaugmented puzzles
from the official Sudoku-Extreme training pool.
We excluded exact matches to puzzles in EqR's released
Sudoku-Extreme training set and potential duplicates of those
puzzles under Sudoku symmetries.
For a fair comparison with EqR, the validation examples were
not added to the training sets.
The test sets were used only for final evaluation.

\paragraph{Datasets for DT.}
For fine-tuning and validation, we used the same $50{,}000$
ID mazes of size $9\times9$ as \citet{schwarzschild2021can}
and \citet{bansal2022endtoend}.
We randomly split these into $40{,}000$ mazes for fine-tuning
and $10{,}000$ for validation.
The 80/20 proportions match those in the original DT implementation.

We generated 100 test mazes at each size $m\times m$,
with $m=9,19,\ldots,99$, using the public code of
\citet{knutson2026logical}.
Each maze was generated using depth-first search, with no
additional edges added to the resulting tree.
Start cells were not restricted to dead ends and could have
degrees 1 through 4.
The target was the unique path from the start cell to the goal cell.

\begin{figure}[t]
\capstart
\begin{center}
\begin{tikzpicture}
\node[inner sep=0] (mazes)
    {\includegraphics[width=\linewidth]{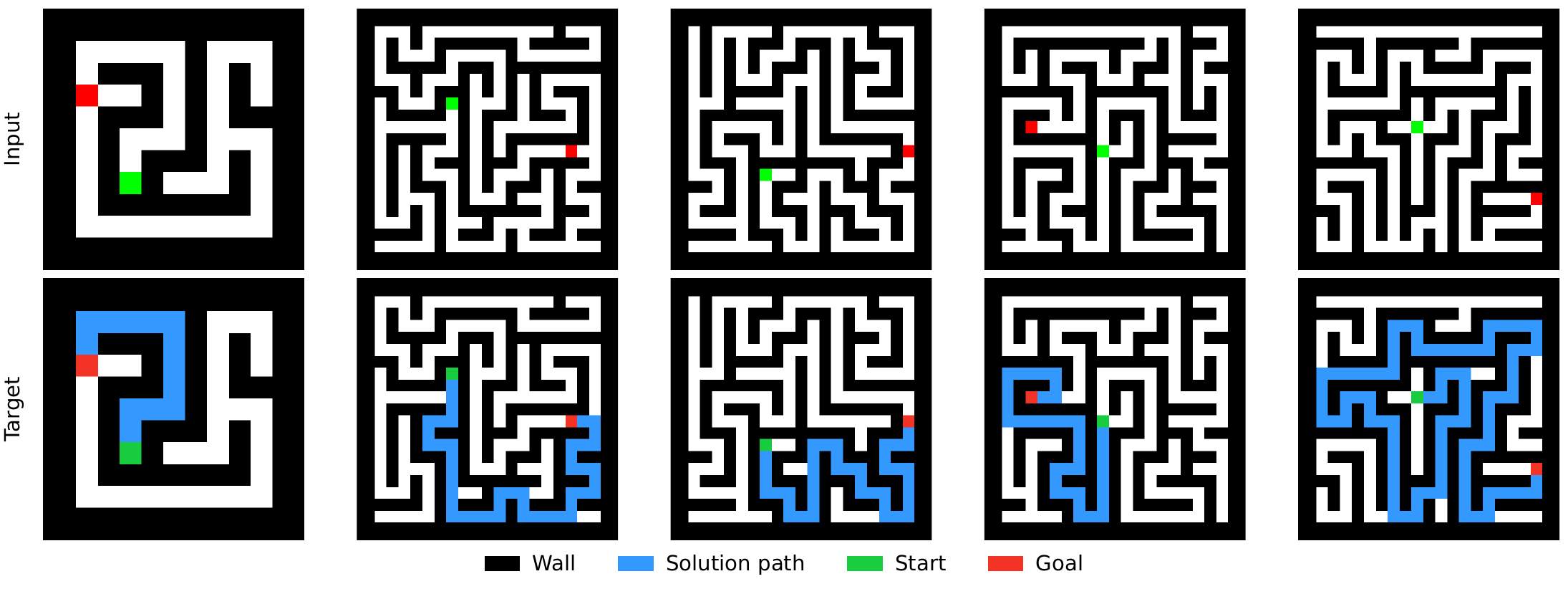}};
\foreach \position/\condition/\size/\degree in
    {0.111/ID/9/1,0.311/OOD/19/1,0.511/OOD/19/2,0.711/OOD/19/3,0.911/OOD/19/4} {
    \node[anchor=south,align=center,yshift=3pt]
        at ($(mazes.north west)!\position!(mazes.north east)$)
        {\condition\ ($\size\times\size$)\\Degree \degree};
}
\end{tikzpicture}
\end{center}
\caption{Examples of ID and OOD mazes used for DT.
The first column shows a $9\times9$ ID validation maze.
The remaining columns show $19\times19$ OOD mazes whose start
cells have degrees 1 through 4.
The top row contains the inputs, and the bottom row marks the
target solution path in blue, the start cell in green, and the
goal cell in red.}
\label{fig:dt-maze-examples}
\end{figure}

Following the input and target format used in prior work
\citep{schwarzschild2021can,bansal2022endtoend},
we converted the generated mazes and solution paths into
input images and target masks.
For a maze of size $m\times m$, the generator used a
$\frac{m+1}{2}\times\frac{m+1}{2}$ grid.
Rendering the passages and walls, including the outer walls,
produced an image of $(m+2)\times(m+2)$ pixels.
We expanded each pixel to $2\times2$ pixels and added
padding of 1 pixel on each side using the pixel value for walls.
The resulting input dimensions were $H=W=2m+6$.
For example, a $9\times9$ maze produced a $24\times24$ input image.
Figure~\ref{fig:dt-maze-examples} shows examples of the ID
validation mazes and OOD test mazes.

Table~\ref{tab:dataset-details} summarizes the dataset sizes
and the counts of generated test mazes by the degree of the
start cell.

\begin{table}[H]
\caption{Dataset sizes (top) and counts of generated test mazes
by the degree of the start cell (bottom).
The degree counts are pooled across all maze sizes.
The Sudoku-Extreme training count includes augmentation.}
\label{tab:dataset-details}
\begin{center}
\begin{tabular}{@{}lrrr@{}}
\toprule
Dataset & Train & Validation & Test \\
\midrule
Maze-Unique
& $1{,}000$ & $1{,}000$ & $1{,}000$ \\
Sudoku-Extreme
& $1{,}001{,}000$ & $10{,}000$ & $422{,}786$ \\
Maze
& $40{,}000$ & $10{,}000$ & $1{,}000$ \\
\bottomrule
\end{tabular}

\medskip

\begin{tabular}{@{}lrrrr@{}}
\toprule
Degree of the start cell & 1 & 2 & 3 & 4 \\
\midrule
Test mazes & 104 & 800 & 95 & 1 \\
\bottomrule
\end{tabular}
\end{center}
\end{table}

\subsection{Model Architectures and the Number of Parameters}
\label{app:models}

We use the model architectures of EqR~\citep{huang2026equilibrium}
and DT~\citep{bansal2022endtoend} unchanged when applying \ImR{}.
Table~\ref{tab:eqr-architecture} summarizes the model architectures
and the number of parameters.

\begin{table}[H]
\caption{Model architectures and the number of parameters.
For EqR, parameters used to embed the input are included
under $f_{\vx}$.
The totals also include the halting head in EqR and the
neural network used to initialize the latent state in DT.}
\label{tab:eqr-architecture}
\begin{center}
\begin{tabular}{@{}lllrrr@{}}
\toprule
Dataset & Position mixing & Puzzle prefix
    & $f_{\vx}$ & $g$ & Total \\
\midrule
Maze-Unique & Self-attention & With
    & $263{,}040$ & $768$ & $264{,}066$ \\
            &                & Without
    & $262{,}912$ & $768$ & $263{,}938$ \\
\midrule
Sudoku-Extreme & MLP & With
    & $5{,}022{,}720$ & $5{,}632$ & $5{,}029{,}378$ \\
               &     & Without
    & $4{,}848{,}640$ & $5{,}632$ & $4{,}855{,}298$ \\
\midrule
Maze & Convolution & Without
    & $740{,}736$ & $39{,}312$ & $783{,}504$ \\
\bottomrule
\end{tabular}
\end{center}
\end{table}

\paragraph{EqR.}
The latent state consists of 2 tensors,
$\vs_t^{(H)},\vs_t^{(L)}\in\mathbb{R}^{(n+p)\times w}$,
where $n$ is the number of grid cells and $w$ is the hidden width.
Here, $p=16$ with the puzzle prefix and $p=0$ without it.
We identify the pair $(\vs_t^{(H)},\vs_t^{(L)})$ with the flattened
state $\vs_t$ in the main text, so $d=2(n+p)w$.
The hidden width is $w=128$ for Maze-Unique and $w=512$
for Sudoku-Extreme.

Let $\mE_{\vx}$ denote the input embedding.
With the puzzle prefix, one learned vector shared across
problems and 15 zero vectors are prepended to the grid embeddings.
The recurrent block $f_{\vx}$ transforms both latent tensors,
including their prefix positions.
Removing the prefix also reduces the input, output, and
intermediate dimensions of the token-mixing MLPs
on Sudoku-Extreme.

Let $B_\theta$ denote the shared network.
It contains 1 transformer block~\citep{vaswani2017attention}
for Maze-Unique and 2 MLP-Mixer blocks~\citep{tolstikhin2021mlp}
for Sudoku-Extreme.
An internal update of a state $h$ with injected input $a$ is
\begin{equation}
    F_\theta(h,a)
    =
    (1-\lambda)h+\lambda B_\theta(h+a),
    \qquad \lambda=0.95.
    \label{eq:eqr-module}
\end{equation}
The recurrent block first updates $\vs^{(L)}$ $J$ times while
holding $\vs^{(H)}$ fixed, and then updates $\vs^{(H)}$ once:
\begin{equation}
\begin{aligned}
    \vu^{(0)}
        &= \vs_t^{(L)}, \\
    \vu^{(j)}
        &= F_\theta\!\left(
            \vu^{(j-1)},\vs_t^{(H)}+\mE_{\vx}
        \right),
        \qquad j=1,\ldots,J, \\
    \vs_{t+1}^{(L)}
        &= \vu^{(J)}, \\
    \vs_{t+1}^{(H)}
        &= F_\theta\!\left(\vs_t^{(H)},\vs_{t+1}^{(L)}\right), \\
    f_{\vx}(\vs_t)
        &= \left(\vs_{t+1}^{(H)},\vs_{t+1}^{(L)}\right).
\end{aligned}
\label{eq:eqr-recurrent-update}
\end{equation}
These $J+1$ internal updates together constitute one application
of $f_{\vx}$ and increase recurrent depth by one.
We use $J=4$ for Maze-Unique and $J=6$ for Sudoku-Extreme,
corresponding to 5 and 14 block applications, respectively.
The parameters of $B_\theta$ are shared across all internal
updates and recurrent depths.
Adding an independent noise sample after each internal update
gives the stochastic map $\tilde{f}_{\vx}(\vs_t,\epsilon_t)$
described in Appendix~\ref{app:stochastic}.

The prediction head reads the grid positions of $\vs_t^{(H)}$.
Let $\vs_{t,\mathrm{grid}}^{(H)}\in\mathbb{R}^{n\times w}$
denote positions ${p+1,\ldots,p+n}$ of $\vs_t^{(H)}$.
The prediction head is
\begin{equation}
    g(\vs_t)
    =
    \vs_{t,\mathrm{grid}}^{(H)}\mW_g^\top
    \in\mathbb{R}^{n\times C},
    \qquad
    \mW_g\in\mathbb{R}^{C\times w}.
    \label{eq:eqr-prediction-head}
\end{equation}
The same linear map is applied at every grid position,
with no bias term.
The halting head reads the first position of $\vs_t^{(H)}$
and uses a weight matrix and a bias to produce 2 outputs.
The first output is used as the halting logit.

The prediction head $g$ defined in
equation~\ref{eq:eqr-prediction-head} is linear, so
\begin{equation}
    \E_{\vs\sim\hat{\mu}_T}[g(\vs)]
    =
    g\!\left(\E_{\vs\sim\hat{\mu}_T}[\vs]\right).
    \label{eq:eqr-head-expectation}
\end{equation}
The halting head also satisfies this identity.
For computational efficiency, we first compute the mean of
$\vs^{(H)}$ over the states in the empirical measure and then
apply each head once.
The prediction head uses the grid positions of this mean,
and the halting head uses its first position.

\paragraph{DT.}
The architecture of DT uses convolutional residual
blocks~\citep{he2016deep}.
As in the main text, the input consists of discrete values
at grid cells.
For the CNN, these discrete values are converted into pixel
values to form an input image.
To make the convolutional operations explicit, we write this
image as $\vx\in\mathbb{R}^{3\times H\times W}$ here.
The latent state is represented as a tensor
$\vs_t\in\mathbb{R}^{128\times H\times W}$.
Flattening this tensor gives the state in the main text,
with $d=128HW$.
A convolution followed by a rectified linear unit (ReLU) activation
initializes the latent state from the input.
The recurrent block then updates this state using input
concatenation, another convolution, and 2 residual blocks:
\begin{equation}
\begin{aligned}
    \vs_0
        &= \operatorname{ReLU}\!\left(C_{\mathrm{in}}(\vx)\right), \\
    f_{\vx}(\vs_t)
        &= B_2\!\left(
            B_1\!\left(C_{\mathrm{rec}}([\vs_t,\vx])\right)
        \right),
\end{aligned}
\label{eq:dt-recurrent-update}
\end{equation}
where $[\vs_t,\vx]$ denotes concatenation along the channel
dimension.
The convolution $C_{\mathrm{in}}$ is used only for initialization,
whereas $C_{\mathrm{rec}}$ is applied at every recurrent depth.
Each residual block has the form
\[
    B_j(\vu)
    =
    \operatorname{ReLU}\!\left(
        \vu+C_{j,2}\!\left(
            \operatorname{ReLU}(C_{j,1}(\vu))
        \right)
    \right),
    \qquad j\in\{1,2\}.
\]
The 2 residual blocks have distinct parameters.
One application of $f_{\vx}$ comprises $C_{\mathrm{rec}}$
and both residual blocks, for a total of 5 convolutions,
and increases recurrent depth by one.
The entire recurrent block shares its parameters across depths.

In tensor form, the prediction head is
\begin{equation}
    g(\vs_t)
    =
    C_{g,3}\!\left(
        \operatorname{ReLU}\!\left(
            C_{g,2}\!\left(
                \operatorname{ReLU}\!\left(C_{g,1}(\vs_t)\right)
            \right)
        \right)
    \right)
    \in\mathbb{R}^{2\times H\times W}.
    \label{eq:dt-prediction-head}
\end{equation}
The convolutions $C_{g,1}$, $C_{g,2}$, and $C_{g,3}$ have
channel dimensions $128\to32$, $32\to8$, and $8\to2$,
respectively.
The 2 output channels contain logits for whether each
pixel belongs to the solution path.
Arranging the pixels as rows and the output channels as columns
gives the logit matrix in the main text, with $n=HW$ and $C=2$
for this image representation.
Since this head is nonlinear, \ImR{} applies it to each state
before computing the empirical expectation of its outputs.

All convolutions use $3\times3$ kernels, stride 1, padding 1,
and no bias.
They preserve the spatial resolution, so the same parameters
can process all evaluated maze sizes.

\subsection{Training and Evaluation Settings}
\label{app:finite-estimation}
\label{app:eqr-segmented-training}

We largely followed the training and evaluation settings of
EqR~\citep{huang2026equilibrium} and
DT~\citep{bansal2022endtoend}.

To reduce the influence of the initial transient, we can exclude
the first $B$ states and construct the empirical measure
\begin{equation}
    \hat{\mu}_{T,B}
    =
    \frac{1}{T-B}\sum_{t=B+1}^{T}\delta_{\vs_t},
    \qquad 0\leq B<T.
    \label{eq:trimmed-estimator}
\end{equation}
The corresponding prediction uses the expectation
$\E_{\vs\sim\hat{\mu}_{T,B}}[g(\vs)]$.

Fine-tuning used trajectories of $T=384$ recurrent updates
with both architectures.
The baseline predicted from $\vs_{384}$, and \ImR{} used
$\hat{\mu}_{384,84}$, which contains the states at depths 85--384.
In the main evaluations, \ImR{} used $B=0$ and therefore
included all states from depth 1 through the stopping depth $T$.
The evaluation depths are given in
Sections~\ref{sec:exp-eqr} and~\ref{sec:exp-dt}.
The validation settings are described below.

Tables~\ref{tab:finetuning-details}
and~\ref{tab:eqr-recurrence-settings} summarize the hyperparameters
for fine-tuning and training from scratch, respectively.
For training from scratch, hyperparameters were selected
separately for each configuration using validation accuracy.
For EqR trained from scratch on Sudoku-Extreme without the
puzzle prefix, $(L,K)=(3,1)$ was selected based on validation
accuracy in a hyperparameter search that also included
$(L,K)=(36,4)$.
The optimizer-update counts in both tables give the full
training budgets.

For all fine-tuning experiments, checkpoint selection included
the initial prediction head before any fine-tuning updates.
Based on validation metrics, fine-tuned heads were selected
for all \ImR{} (FT) experiments and for EqR (FT) on Maze-Unique,
whereas the initial head was retained for EqR (FT) on Sudoku-Extreme.

\begin{table}[H]
\caption{Hyperparameters for fine-tuning the prediction head.}
\label{tab:finetuning-details}
\begin{center}
\begin{tabular}{@{}llrr@{}}
\toprule
Architecture & Dataset
    & \shortstack{Batch\\size}
    & \shortstack{Optimizer\\updates} \\
\midrule
EqR & Maze-Unique     & 768 & $1{,}000$ \\
EqR & Sudoku-Extreme & 768 & $1{,}000$ \\
DT  & Maze           & 50  & $1{,}600$ \\
\bottomrule
\end{tabular}
\end{center}
\end{table}

\begin{table}[H]
\caption{Hyperparameters selected for training from scratch
with the architecture of EqR.
Here, $L$ is the number of recurrent updates per inner loop,
and gradients through the recurrent block are retained
for the final $K$ updates.}
\label{tab:eqr-recurrence-settings}
\begin{center}
\begin{tabular}{@{}lllrrrr@{}}
\toprule
Model & Dataset & Puzzle prefix & $L$ & $K$
    & \shortstack{Batch\\size}
    & \shortstack{Optimizer\\updates} \\
\midrule
\ImR{} & Maze-Unique    & With / Without & 36 & 4
    & 768 & $150{,}000$ \\
\ImR{} & Sudoku-Extreme & With / Without & 36 & 4
    & 768 & $50{,}000$ \\
EqR & Maze-Unique       & Without        & 36 & 4
    & 768 & $150{,}000$ \\
EqR & Sudoku-Extreme    & Without        & 3  & 1
    & 768 & $50{,}000$ \\
\bottomrule
\end{tabular}
\end{center}
\end{table}

\paragraph{EqR initialization and noise.}
When a new example was loaded, the entries of $\vs^{(H)}$ and
$\vs^{(L)}$ were sampled independently from a standard normal
distribution truncated to $[-2,2]$ and rescaled so that
the resulting distribution had standard deviation 1.

During training and validation, each internal update
$F_\theta$ added an independent Gaussian noise sample
with standard deviation 0.01.
The noise was added after the weighted combination in
equation~\ref{eq:eqr-module}, including every update of
$\vs^{(L)}$ and $\vs^{(H)}$.
For the main evaluations, we set this noise to zero.

\paragraph{EqR training from scratch.}
Training followed the deep supervision procedure in
Section~\ref{sec:atr-training}.
We used AdamATan2~\citep{everett2024scaling} with
$(\beta_1,\beta_2)=(0.9,0.95)$ and weight decay 1.0
for all parameters except the shared prefix vector.
The learning rate increased linearly to $10^{-4}$ over
the first $2{,}000$ optimizer updates and remained constant
thereafter.
The shared prefix vector was updated separately using
signSGD~\citep{bernstein2018signsgd} with learning rate
$10^{-4}$ and weight decay 1.0.

Following EqR, we maintained an exponential moving average
(EMA)~\citep{song2020improved}, with coefficient 0.999,
of the parameters updated by AdamATan2.
We used these EMA parameters for validation and evaluation.
The prefix vector was used without EMA.

We used EqR's original task loss, averaged over valid tokens
within each example.
We added the binary cross-entropy loss for the halting
prediction with weight 0.5 and averaged the combined loss
over the batch.
The binary target for halting indicated whether the current
prediction matched the entire target grid.

Each example was processed for at most 16 outer steps.
The outer loop stopped for an example when its halting logit
was positive, except when the following rule required
computation to continue.
At each outer step, with probability 0.1, we sampled
a minimum outer-step count uniformly from 2 through 16.
For that step, we ignored a positive halting logit if
the current count was below the sampled minimum.
Completed examples were replaced by new examples with
newly sampled latent states.

Validation used 16 outer steps for each example.
Predictions were computed from the final outer step:
\ImR{} used the expectation under the empirical measure
of all $L$ states in its inner loop, and EqR predicted
from the final state of that loop.
For evaluation, we used the model weights from the point
in training with the highest accuracy on the validation set.

\paragraph{EqR fine-tuning.}
For EqR with the puzzle prefix, we used the released weights
for evaluation and to initialize fine-tuning.
We initialized a new AdamATan2 optimizer for the prediction
head, with learning rate $10^{-4}$,
$(\beta_1,\beta_2)=(0.9,0.95)$, and weight decay 1.0.
The learning rate remained constant throughout fine-tuning.
Each optimizer update used a new batch and newly sampled
latent initial states.

Validation used trajectories of 384 recurrent updates.
EqR (FT) predicted from $\vs_{384}$, and
\ImR{} (FT) used the expectation under
$\hat{\mu}_{384,0}$, which contains all states at
depths 1--384.
We used accuracy on the validation set to choose how many
fine-tuning updates to use for the prediction head.
For Sudoku-Extreme, EqR fine-tuning did not improve
validation accuracy, so the original prediction head
was retained.
We therefore do not show a separate fine-tuned EqR curve
in Figure~\ref{fig:eqr-main}.

Fine-tuned prediction heads were used without EMA.
All other model parameters remained at their released values.

\paragraph{DT fine-tuning.}
Both fine-tuning methods started from the same released
model weights and the corresponding Adam optimizer state.
We used Adam~\citep{kingma2015adam} with a constant learning
rate of $10^{-4}$, $(\beta_1,\beta_2)=(0.9,0.999)$,
and weight decay $2\times10^{-4}$.
The task loss was cross-entropy computed only on traversable
pixels.

Validation ran to recurrent depth 384 for both methods.
DT (FT) used the logits $g(\vs_{384})$.
For \ImR{} (FT), the validation logits were
\[
    \E_{\vs\sim\hat{\mu}_{384,0}}[g(\vs)]
    =
    \frac{1}{384}\sum_{t=1}^{384}g(\vs_t).
\]
The prediction head was applied to each state before computing
this expectation, and the initial state $\vs_0$ was not included.
For evaluation, we used the prediction-head weights from
the point in fine-tuning with the lowest loss
on the ID validation set.

\subsection{Classification of Latent Trajectories}
\label{app:eqr-fixed-diagnostics}

We classified the latent trajectories reported in
Table~\ref{tab:eqr-from-scratch-dynamics} using changes in
$\vs_{t,\mathrm{grid}}^{(H)}$, the grid positions of $\vs_t^{(H)}$
used by the prediction head, as described in
Appendix~\ref{app:models}.
After each recurrent update, we computed the relative
$L_\infty$ residual
\begin{equation}
    r_t
    =
    \frac{
        \lVert
        \vs_{t+1,\mathrm{grid}}^{(H)}
        -
        \vs_{t,\mathrm{grid}}^{(H)}
        \rVert_\infty
    }{
        \lVert\vs_{t+1,\mathrm{grid}}^{(H)}\rVert_\infty
        +10^{-8}
    }.
    \label{eq:eqr-relative-residual}
\end{equation}
Here, $\lVert\cdot\rVert_\infty$ denotes the largest absolute
entry across all grid positions and latent dimensions.

We evaluated trajectories to depth $T=512$ on Maze-Unique
and $T=1024$ on Sudoku-Extreme.
A trajectory was classified as fixed if
\begin{equation}
    \max_{t=T-64,\ldots,T-1} r_t < 0.05,
    \label{eq:eqr-fixed-criterion}
\end{equation}
and as non-fixed otherwise.
Thus, the fixed label requires small relative changes
over all of the final 64 recurrent updates.
This diagnostic does not guarantee convergence of the
full latent state to a fixed point.

We computed the proportion of non-fixed trajectories
among all evaluated examples.
For prediction stability, we applied the categories
defined in Section~\ref{sec:experiments} over depths
$T-63,\ldots,T$.
The proportion of each prediction category was computed
separately within the fixed and non-fixed groups,
with the number of examples in that group as the denominator.

For each configuration trained or fine-tuned in this study,
we computed the proportions separately for each of the
3 training seeds and reported their mean and standard
deviation.
For EqR with the puzzle prefix, we reported results
from the single released model for each dataset.
Fine-tuning updates only the prediction head, so the
proportions of fixed and non-fixed trajectories remain
unchanged.

\subsection{Visualization of Latent Trajectories and Predictions}
\label{app:latent-dynamics-visualization}

\begin{figure*}[t]
\centering
\includegraphics[width=\textwidth]{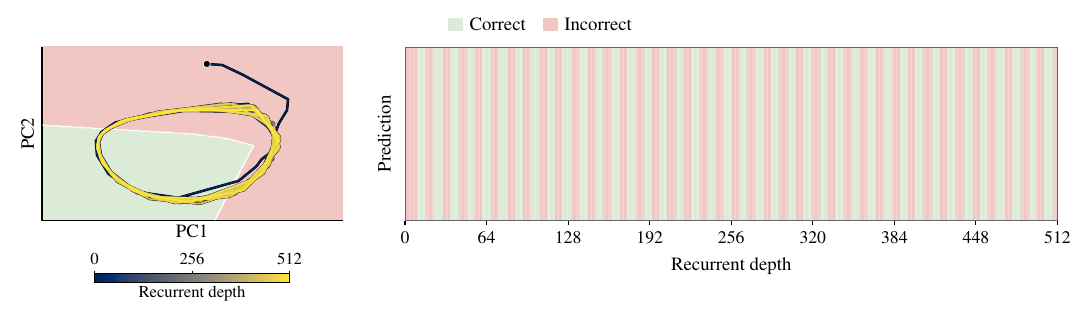}
\caption{The EqR trajectory in Figure~\ref{fig:latent-reasoning-overview}(a)
and its actual predictions.
Left: the trajectory over depths 0--512 and prediction regions
on the reconstructed PCA plane.
Trajectory color indicates recurrent depth; the circle marks
the initial state.
Right: whole-maze correctness from the original head applied
to each actual latent state.
Pale green and pale red indicate correct and incorrect
predictions, respectively.}
\label{fig:eqr-trajectory-actual-predictions}
\end{figure*}

For Figure~\ref{fig:latent-reasoning-overview}(a), we used the publicly released
EqR model for Maze-Unique with its EMA parameters, original
prediction head, and puzzle prefix. For one test maze, we ran
512 recurrent updates from the model's standard random
initialization with zero transition noise.
Figure~\ref{fig:eqr-trajectory-actual-predictions} reproduces this
trajectory alongside the predictions from its actual latent states.

We fitted PCA to the same 513 states shown in the figure,
including the initial state at depth~0. Let
$\vu_t=\operatorname{vec}(\vs_{t,\mathrm{grid}}^{(H)})$ denote
the flattened high-level state excluding the puzzle prefix.
With mean $\bar{\vu}$ and the first two principal directions
$\mV$, the displayed trajectory is
\[
    \vq_t=\mV^\top(\vu_t-\bar{\vu}),
    \qquad t=0,\ldots,512.
\]
These components explain 77.0\% of the variance.
The trajectory was generated by the full recurrent model;
PCA was applied only for visualization.

For the background, we evaluated the original prediction head
on a $400\times260$ grid of reconstructed states
\[
    \vu(\vq)=\bar{\vu}+\mV\vq,
\]
with all omitted PCA components fixed at their mean.
A point is correct only if all 900 predicted maze cells match
the target. The white line separates the correct and incorrect
regions. This boundary is defined on a two-dimensional slice
and only approximately represents the prediction regions
visited by the full trajectory: omitted components can change
the prediction at the same projected location.

The right panel of Figure~\ref{fig:eqr-trajectory-actual-predictions}
reports the original head's predictions at the actual latent
states, without PCA reconstruction. These predictions repeatedly
alternate between correct and incorrect. Thus, the original
high-dimensional trajectory also repeatedly visits both prediction
regions, supporting the qualitative interpretation of
Figure~\ref{fig:latent-reasoning-overview}(a). The precise geometry of the
high-dimensional boundaries, however, cannot be inferred from
the two-dimensional visualization alone.

\section{Additional Results}
\label{app:additional-results}

This appendix presents accuracy at the final recurrent depth,
PCA projections of latent trajectories, and DT results
by maze size.

\subsection{Accuracy at the Final Recurrent Depth}
\label{app:final-depth-accuracy}

Table~\ref{tab:final-depth-accuracy} reports accuracy at the
final recurrent depths used in the EqR and DT experiments.
For DT, results are grouped by the degree of the start cell,
as in Section~\ref{sec:exp-dt}.

For both prefix settings, \ImR{} trained from scratch
achieves higher accuracy than EqR on both datasets.
Fine-tuning with \ImR{} yields higher accuracy than
the fine-tuned baseline on Maze-Unique and both groups
of DT mazes, but lower accuracy on Sudoku-Extreme.

\begin{table}[H]
\caption{Accuracy at the final recurrent depth.
Entries with $\pm$ report the mean and standard deviation
across 3 training seeds.
The remaining entries use a single released model.}
\label{tab:final-depth-accuracy}
\begin{center}
\begin{tabular}{@{}llr@{}}
\toprule
Model & Puzzle prefix & Accuracy (\%) \\
\midrule
\multicolumn{3}{@{}l}{\emph{Maze-Unique}, $T=512$} \\
EqR                    & With    & $89.4$ \\
EqR (FT)        & With    & $89.4 \pm 0.0$ \\
\ImR{} (FT)     & With    & $91.6 \pm 0.1$ \\
\ImR{}                 & With    & $98.8 \pm 0.9$ \\
\addlinespace[2pt]
EqR                    & Without & $82.9 \pm 20.9$ \\
\ImR{}                 & Without & $99.0 \pm 1.4$ \\
\midrule
\multicolumn{3}{@{}l}{\emph{Sudoku-Extreme}, $T=1024$} \\
EqR                    & With    & $96.2$ \\
EqR (FT)        & With    & $96.2 \pm 0.0$ \\
\ImR{} (FT)     & With    & $95.4 \pm 0.0$ \\
\ImR{}                 & With    & $98.7 \pm 0.2$ \\
\addlinespace[2pt]
EqR                    & Without & $96.2 \pm 0.3$ \\
\ImR{}                 & Without & $99.0 \pm 0.2$ \\
\midrule
\multicolumn{3}{@{}l}{
    \emph{OOD mazes, degree of the start cell: 2},
    $T=5{,}000$} \\
DT                     & -- & $77.75$ \\
DT (FT)         & -- & $82.83 \pm 0.14$ \\
\ImR{} (FT)     & -- & $84.50 \pm 0.00$ \\
\midrule
\multicolumn{3}{@{}l}{
    \emph{OOD mazes, degree of the start cell: 3},
    $T=5{,}000$} \\
DT                     & -- & $25.26$ \\
DT (FT)         & -- & $26.32 \pm 0.00$ \\
\ImR{} (FT)     & -- & $28.42 \pm 0.00$ \\
\bottomrule
\end{tabular}
\end{center}
\end{table}

\subsection{Projections of Latent Trajectories}
\label{app:eqr-pca}

\begin{figure}[t]
\capstart
\begin{center}
\includegraphics[width=\linewidth]{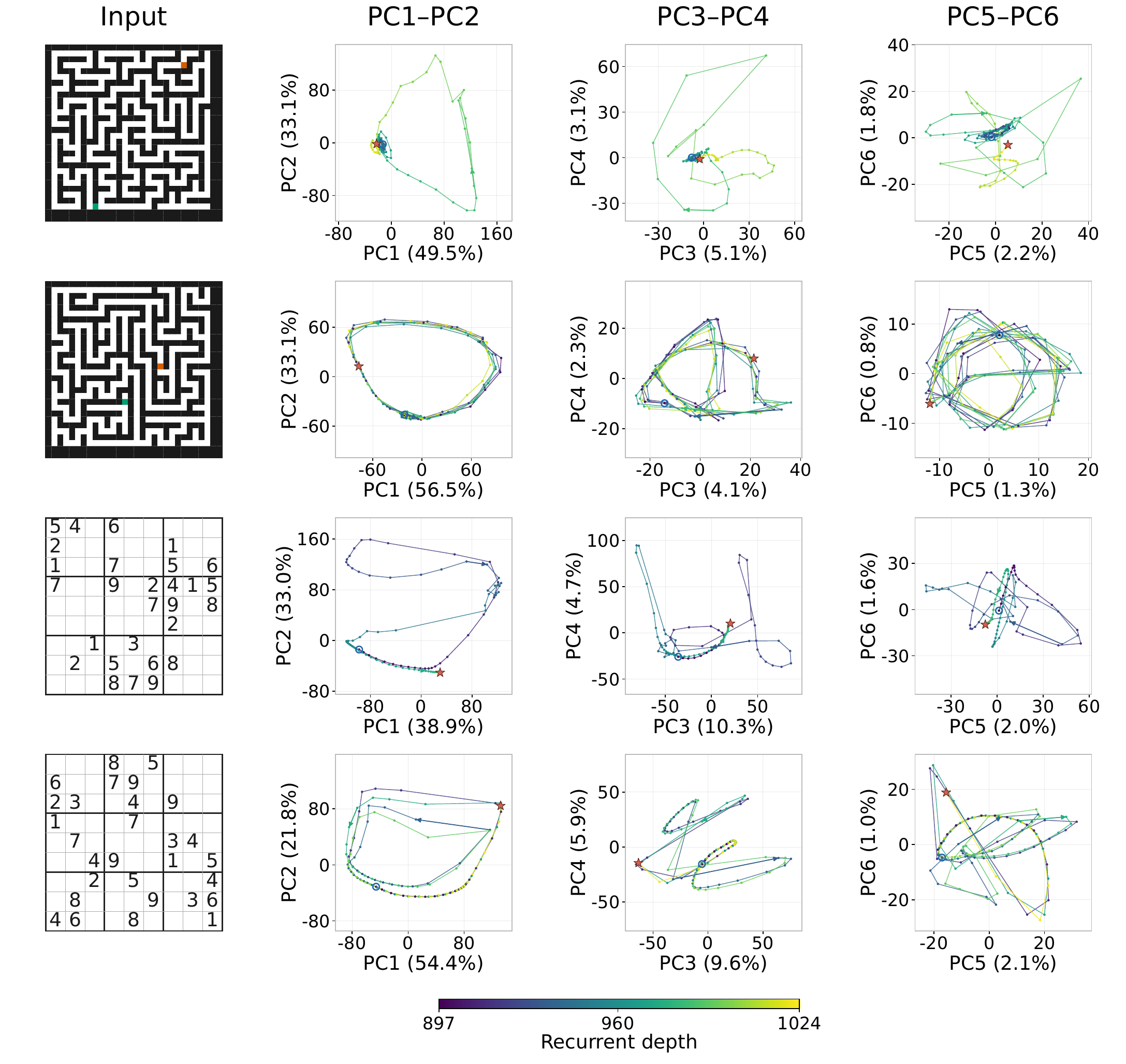}
\end{center}
\caption{PCA projections of the part of the latent state
used by the prediction head for \ImR{} models trained
without the puzzle prefix.
Each row shows an input and the corresponding latent
trajectory projected onto PC1--PC2, PC3--PC4, and PC5--PC6.
The first two rows show Maze-Unique examples,
and the last two show Sudoku-Extreme examples.
Axis labels give the percentage of variance explained
by each principal component.
Colors indicate recurrent depths 897--1024;
circles and stars mark the first and last states
in this interval, respectively.
All trajectories were generated with noise set to zero.}
\label{fig:eqr-nonfixed-pca}
\end{figure}

Figure~\ref{fig:eqr-nonfixed-pca} shows PCA projections of
$\vs_{t,\mathrm{grid}}^{(H)}$, the part of the latent state
used by the prediction head.
For these visualizations, we display depths 897--1024
for both Maze-Unique and Sudoku-Extreme.
The trajectories show continued state changes over
this interval with noise set to zero.

\subsection{DT Results by Maze Size}
\label{app:dt-full-results}

\begin{figure}[t]
\capstart
\begin{center}
\includegraphics[
        width=\linewidth,
        height=0.80\textheight,
        keepaspectratio
    ]{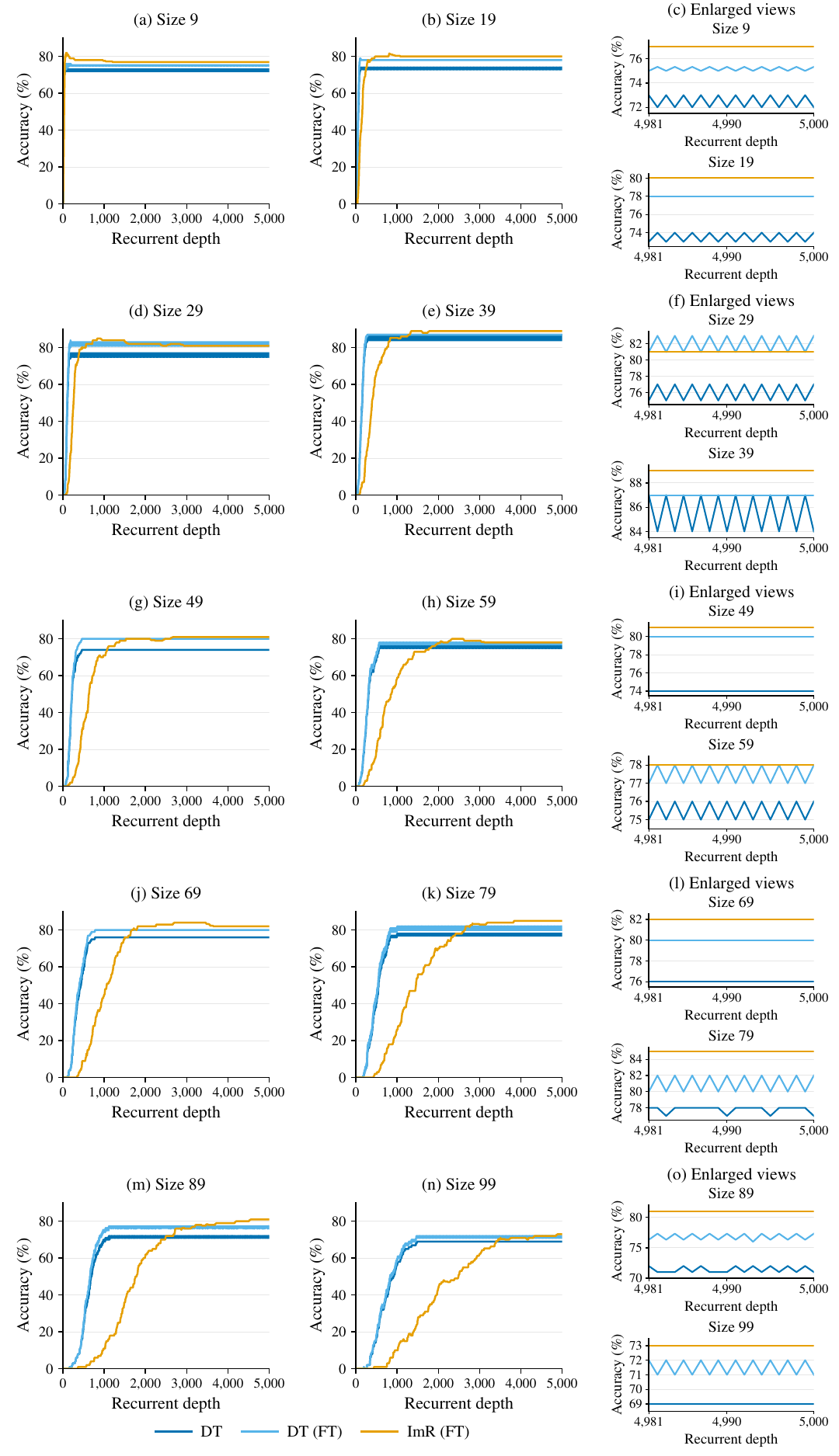}
\end{center}
\caption{Accuracy across recurrent depth for each maze size.
The first two columns show sizes $9\times9$ through
$99\times99$, with 100 test mazes per size,
combining all degrees of the start cell.
The right column shows enlarged views over
depths $4{,}981$--$5{,}000$.
Curves for DT (FT) and \ImR{} (FT)
report means across 3 training seeds.}
\label{fig:dt-size-depth-accuracy}
\end{figure}

Figure~\ref{fig:dt-size-depth-accuracy} compares accuracy
across recurrent depth separately for each maze size.
The enlarged views illustrate the reduction in accuracy
fluctuations at large recurrent depths with
\ImR{} (FT).

\end{document}